\documentclass[a4paper,11pt]{article}
\usepackage[T1]{fontenc}
\usepackage[utf8]{inputenc}
\usepackage{amsmath}
\usepackage[margin=1.0in]{geometry}
\usepackage{graphicx}
\usepackage{gensymb}
\usepackage[numbers,sort&compress]{natbib}
\usepackage{rotating}
\usepackage{xcolor}
\usepackage{colortbl}
\usepackage{placeins}
\usepackage{array}
\usepackage{authblk}
\usepackage{footnote}
\makesavenoteenv{table}
  {\par\vspace{\abovedisplayskip}\noindent\begin{tabular}{>{$}l<{$} @{${}={}$} l}}
  {\end{tabular}\par\vspace{\belowdisplayskip}}

\newcommand{\grayline}{\arrayrulecolor{lightgray}\hline\arrayrulecolor{black}}   

\newcommand{\keywords}[1]{%
  \begin{center}
    \begin{minipage}{0.9\textwidth}
      \small\textbf{Keywords:} #1
    \end{minipage}
  \end{center}
}

\begin{document}
\title{Inclusion of obstacles in ground point classification in ALS data for terrain models used for analysis of vehicle traversability}
\author[1]{Mikael Lundb\"{a}ck}
\author[1]{Erik Wallin\thanks{Corresponding author: erik.wallin@umu.se}}
\author[2]{Sima Mohtasami}
\author[1]{Martin Servin}
\author[3]{Mattias Nystr\"{o}m}
\author[2,4]{Henrik Persson}
\affil[1]{Department of Physics, Ume\aa\ University, Ume\aa, Sweden}
\affil[2]{Skogforsk, Uppsala, Sweden}
\affil[3]{Komatsu Forest AB, Ume\aa, Sweden}
\affil[4]{Department of Forest Resource Management, Swedish University of Agricultural Sciences, Uppsala, Sweden}
\date{}

\maketitle

\begin{abstract}
Classification of ALS point clouds has been studied for decades, but it remains challenging in forest contexts due to the great variety of terrain types, sensor equipment, flight parameters, and resulting point densities.
Point cloud density has increased significantly in recent years, and existing algorithms, developed for lower density point clouds, may no longer be effective.
It is beyond the scope of most existing algorithms to classify obstacles, such as boulders, while also distinguishing non-obstacles, such as bushes and undergrowth.
We evaluate the performance of different algorithms for ground point classification in ALS data in a context where obstacles are included in the ground class, resulting in terrain models useful for offroad machinery route planning and navigation.
The evaluation is performed using metrics directly on the point clouds (Precision, Recall, $F_1$-score) and on resulting terrain models (RMSE and maps of differences).
The results show that there are several algorithms that perform well, but performance is highly affected by the parameters used.
The best results in terms of terrain model RMSE are obtained with the CSF-algorithm, with an average RMSE of 0.08 meters.
The results indicate that existing algorithms can be adapted to include obstacles in the ground class under the tested conditions.
\end{abstract}

\keywords{airborne laser scanning, digital terrain model, terrain trafficability, forest operations, off-road navigation}

\section{Introduction}
Accurate classification of ground points from Aerial Laser Scanning (ALS) point clouds is crucial for creating reliable terrain models, which in turn are important for route planning and navigation of heavy machinery in rough terrain.
With increasing point cloud density, earlier works and algorithms developed for lower density point clouds may need critical evaluation.
In forestry applications, ALS is particularly valuable for its ability to penetrate the canopy and capture detailed information about the forest floor.
A core step in producing a reliable Digital Terrain Model (DTM) from LiDAR data is the classification of ground points, which is done in different ways with different algorithms \citep{chen2017state}.
However, it is a challenge to classify points in a way that identifies the ground together with obstacles, such as boulders, while distinguishing non-obstacles, like bushes and undergrowth \citep{puliti_for-instance_2023}.

Conventional DTMs are used for a number of applications, such as reference for tree heights in forestry, soil moisture estimations, calculation of slope and aspect, and visualization of the terrain surface.
For these applications, so called bare ground DTMs are preferred and used, often with LiDAR point clouds as data source \citep{lohani2017airborne}.
The common goal of all these algorithms is to separate points belonging to bare ground from points belonging to other objects, such as trees, bushes, and boulders.
With the increased use of high-density LiDAR data, the need for different algorithms for ground point classification has increased, as the increased point density makes it possible to capture more detailed terrain structures.

Detailed terrain models are useful for analysis of machine-terrain interactions in rough terrain prior to, or during harvesting and other forestry operations, where it is important to know the exact location of obstacles such as boulders to analyse the possibility of the machine to traverse the terrain and estimate its energy consumption.
Apart from route planning, detailed terrain models can also be beneficial in operational navigation of autonomous and tele-operated machines, where operators lack much of the visual, audio, and haptic feedback from the machine-terrain interaction.

In forestry applications, LiDAR sensors are commonly mounted on drones, helicopters, or airplanes to scan while flying in regular patterns at suitable altitudes \citep{lohani2017airborne}.
Factors as the altitude above the ground and the overlap between flight paths influence the resulting point cloud density and thus the ability to accurately reconstruct the ground under the tree canopy \citep{liu_effect_2007}.
The capability to penetrate \emph{through} the canopy is determined e.g. by the laser wavelength, beam divergence, and pulse output rate with wavelengths in the near-infrared range being a good choice for this purpose.
Typical beam divergence is on the order of milliradians, and typical pulse output rates range from $10^5$ to $10^6$ pulses per second.
The number of returns per pulse can be used to distinguish between ground and non-ground points, as the ground typically has a higher number of returns than other objects \citep{zhang2003progressive}.

There are several types of algorithms for ground point classification in ALS data, such as triangular irregular network \citep{axelsson2000generation}, progressive morphological filter \citep{zhang2003progressive}, cloth simulation filter \citep{zhang2016easy}, threshold-free classification \citep{bartels2010threshold}, and multiscale curvature classification \citep{evans2007multiscale}.
In recent years, machine learning approaches to ground classification have become more common \citep{jin2020point,hu2016deep}, as well as methods for end-to-end classification of complete point clouds \citep{chen2021pointnet++}.

Previous works have compared different algorithms for ground point classification in ALS LiDAR data in general \citep{klapvstve2020sensitivity,silva2018comparing,tinkham2011comparison}.
For forested areas, the ground classification is mostly done using algorithms that seek bare ground, to be used as reference for tree heights.
Some efforts have been made to investigate the use of ALS-generated terrain models for terrain trafficability \citep{niemi2017airborne}, and related work on obstacle detection or segmentation typically treats obstacles as a separate target rather than as part of the ground surface representation.
However, for heavy machinery route planning and navigation, it is of relevance to include obstacles that the machinery needs to navigate in the ground class itself.
To the best of our knowledge, we are not aware of previous studies in forested areas that explicitly evaluate existing ALS ground-classification algorithms for this obstacle-inclusive ground definition.
We therefore explore parameter modifications for existing algorithms to investigate their effect on the classification results, and compare the results to manually annotated ground truth data.

\subsection{Aims and objectives}
The aim of this study is to evaluate the performance of different algorithms for ground point classification in ALS data in a context where obstacles like boulders are included in the ground class, resulting in terrain models used for analysis of off-road vehicle traversability.
The objectives of the study are to:
\begin{itemize}
    \item Evaluate the performance of different classification algorithms for ground point classification in ALS data, including obstacles in the ground class.
    \item Investigate the effect of different algorithm parameters on the classification results.
    \item Investigate the effect of different point densities and terrain conditions on the classification results.
    \item Compare the computational speed of a number of implemented algorithms.
\end{itemize}

\section{Materials and methods}
\begin{figure}[h]
    \centering
    \includegraphics[width=0.95\textwidth]{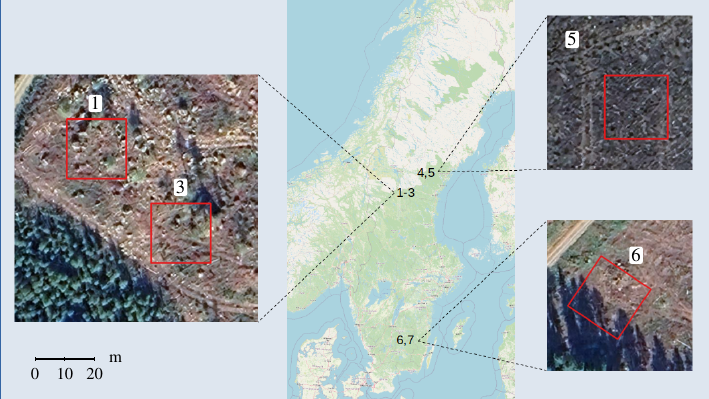}
    \caption{Map of the included field plots, from north to south 'SCA Laxsjö' testsite (plot 4-5), 'Märrviken' testsite (plot 1-3), and 'SLU Asa' research park (plot 6-7).}
    \label{fig:map}
\end{figure}

Seven field plots ($20 \times $20 m) from different locations in Sweden are used in this study (Figure \ref{fig:map}).
The study plots are all scanned with high or very high resolution airborne laser (ALS), and have different terrain conditions.
We investigate different algorithms for ground classification, with different parameter settings, compute a digital terrain model, or raster/heightfield as we interchangeably name it in the following, from the classified points in each setting, and compare the results to a raster produced from manually annotated ground truth.
The comparisons are made in terms of Root Mean Square Error (RMSE) of heights and visual inspection of map differences.
We also compare the point clouds classified automatically by the algorithms to the manually classified point clouds directly.
The evaluation is done using metrics such as Precision, Recall, and $F_1$-score.

\subsection{Data}
We use ALS data from three different sites (Figure \ref{fig:map}), each with different point densities collected over different terrain conditions (Table \ref{tab:sites}).
All datasets were collected in mature forests before clear-cutting but originate from different campaigns with different sensors and flight parameters (Table \ref{tab:sites}).
This heterogeneity is useful for testing practical robustness across realistic acquisition conditions, but it also means that point density is not isolated and instead covaries with sensor characteristics, acquisition geometry, and terrain and vegetation conditions.

\begin{sidewaystable}[h]
    \centering
    \caption{Datasets included in the study. Point density is calculated as the total number of points divided by the area of each field plot, i.e. including multiple returns. Roughness is set according to the Swedish terrain classification scheme, where 5 = very rough terrain and 1 = very smooth terrain \citep{berg1992terrain}. Field measurements were made during autumn 2023 and 2024 and included counting and measurements of obstacles.}
    \label{tab:sites}
    \scriptsize
    \begin{tabular}{p{0.9cm} p{0.9cm} p{2cm} p{2.2cm} p{1.9cm} p{2cm} p{3.5cm} p{2cm} p{2cm} p{3cm}}
        \hline
        \textbf{Site} & \textbf{Field plot} & \textbf{Point density (\ensuremath{\text{points/m}^2})} & \textbf{Wavelengths (nm)} & \textbf{Beam divergence (mrad)} & \textbf{Flight height (m)} & \textbf{Scanner} & \textbf{Roughness (1-5)} & \textbf{Slope\footnote{Average slope of field plot measured in pointcloud} (\degree)} & \textbf{Comments} \\
        \hline
        1 & 1 & 2500 & 1550 & 0.5 & 150 & Riegl VUX 160 & 5\footnote{\label{first}measured in the field} & 2 & big boulders \\
        1 & 2 & 1300 & 1550 & 0.5 & 150 & Riegl VUX 160 & 4\footref{first} & 8 & \\
        1 & 3 & 1800 & 1550 & 0.5 & 150 & Riegl VUX 160 & 4\footref{first} & 1 & very big boulder \\
        2 & 4 & 1100 & 1550 & 0.5 & 70 & Riegl LMS-Q680i & 1\footnote{\label{second}estimated from remote sensing data} & 16 & slight 'shelf' \\
        2 & 5 & 1500 & 1550 & 0.5 & 70 & Riegl LMS-Q680i & 1\footref{second} & 1 & flat \\
        3 & 6 & 70 & 532/1064 & 0.7/0.2 & 800 & Riegl VQ-1560i & 3\footref{first} & 1 & flat \\
        3 & 7 & 35 & 532/1064 & 0.7/0.2 & 800 & Riegl VQ-1560i & 5\footref{first} & 7 &  \\
        \hline
    \end{tabular}
\end{sidewaystable}

\subsection{Manual classification}
As a starting point for the ground truth used for evaluation, we use Terrascan with very aggressive settings (high ``iteration-angle'' and ``iteration-distance'') (Table \ref{tab:terrasettings}) to include boulders as ground points while excluding e.g. trees.
The reference used in the evaluation is thus a manual refinement of a Terrascan-derived initialization rather than a fully independent annotation from scratch, which may favor Terrascan in point-to-point metrics.
Iteration angle decides the angle between a point and its projection on a preliminary ground surface, within which the point is considered to be part of the ground \citep{terrasolid:2024}.
Iteration distance makes sure that the iteration does not make big jumps upward if triangles are large. This avoids ground points that are too high, for example within low vegetation or on low buildings \citep{terrasolid:2024}.
To adjust misclassified ground points we do a manual classification in the software CloudCompare \citep{cloudcompare:2024}.
The software allows the user to manually select groups of points in the point cloud and assign them to a desired class.
The classes stored in the point clouds are ground (2) and unclassified (1), but for the purposes of this study all points not belonging to the ground class are treated analytically as non-ground.
The ground class is assigned to points that are part of the ground surface or obstacles such as boulders, while the non-ground class is assigned to all other points.
The manual classification focuses on removing misclassified points such as trees and bushes from the ground class.
However, unclassified points clearly representing ground or obstacles are also corrected.
The classification was done in a 3D view, where the user could rotate and zoom the view as well as subset the point cloud to get a better understanding of the points.
The manual classification was performed by a single annotator in CloudCompare. The work focused on correcting obvious misclassifications, primarily vegetation incorrectly included in the ground class and omitted ground or obstacle points. No formal annotation agreement analysis was performed.
As visual support, we use photogrammetric point clouds produced from drone imagery captured after harvesting for plots 1-5, and results from a field inventory of boulders and stumps for plots 6-7.
The proportion of ground classified points in the manually classified point clouds ranged from 20 to 77\% (Table \ref{tab:groundshare}).

\begin{table}[h]
    \centering
    \caption{Terrascan settings used for manual classification}
    \label{tab:terrasettings}
    \scriptsize
    \begin{tabular}{p{5cm}p{5cm}}
        \hline
        \textbf{Parameter} & \textbf{Value} \\
        \hline
        \textbf{Max. building size} & 10 m \\
        \textbf{Terrain angle} & 88\degree \\
        \textbf{Iteration angle} & 60\degree  to plane \\
        \textbf{Iteration distance} & 0.9 m to plane \\
        \textbf{Reduce iteration angle} & edge length $<$ 0.3 m \\
        \hline
    \end{tabular}
\end{table}

\begin{table}[h]
    \centering
    \caption{The proportion of ground points in the manually annotated point clouds}
    \label{tab:groundshare}
    \scriptsize
    \begin{tabular}{p{0.9cm} p{0.9cm} p{2cm} p{2.0cm} p{2.0cm} }
        \hline
        \textbf{Site} & \textbf{Field plot} & \textbf{Total (\ensuremath{\text{points/m}^2})} & \textbf{Ground (\ensuremath{\text{points/m}^2})} & \textbf{Proportion groundpoints (\%)} \\
        \hline
        1 & 1 & 2500 & 620 & 25 \\
        1 & 2 & 1300 & 476 & 37 \\
        1 & 3 & 1800 & 446 & 25 \\
        2 & 4 & 1100 & 158 & 14 \\
        2 & 5 & 1500 & 298 & 20 \\
        3 & 6 & 70 & 30 & 43 \\
        3 & 7 & 35 & 27 & 77 \\
        \hline
    \end{tabular}
\end{table}

\begin{figure}[h]
    \centering
    \includegraphics[width=0.9\textwidth]{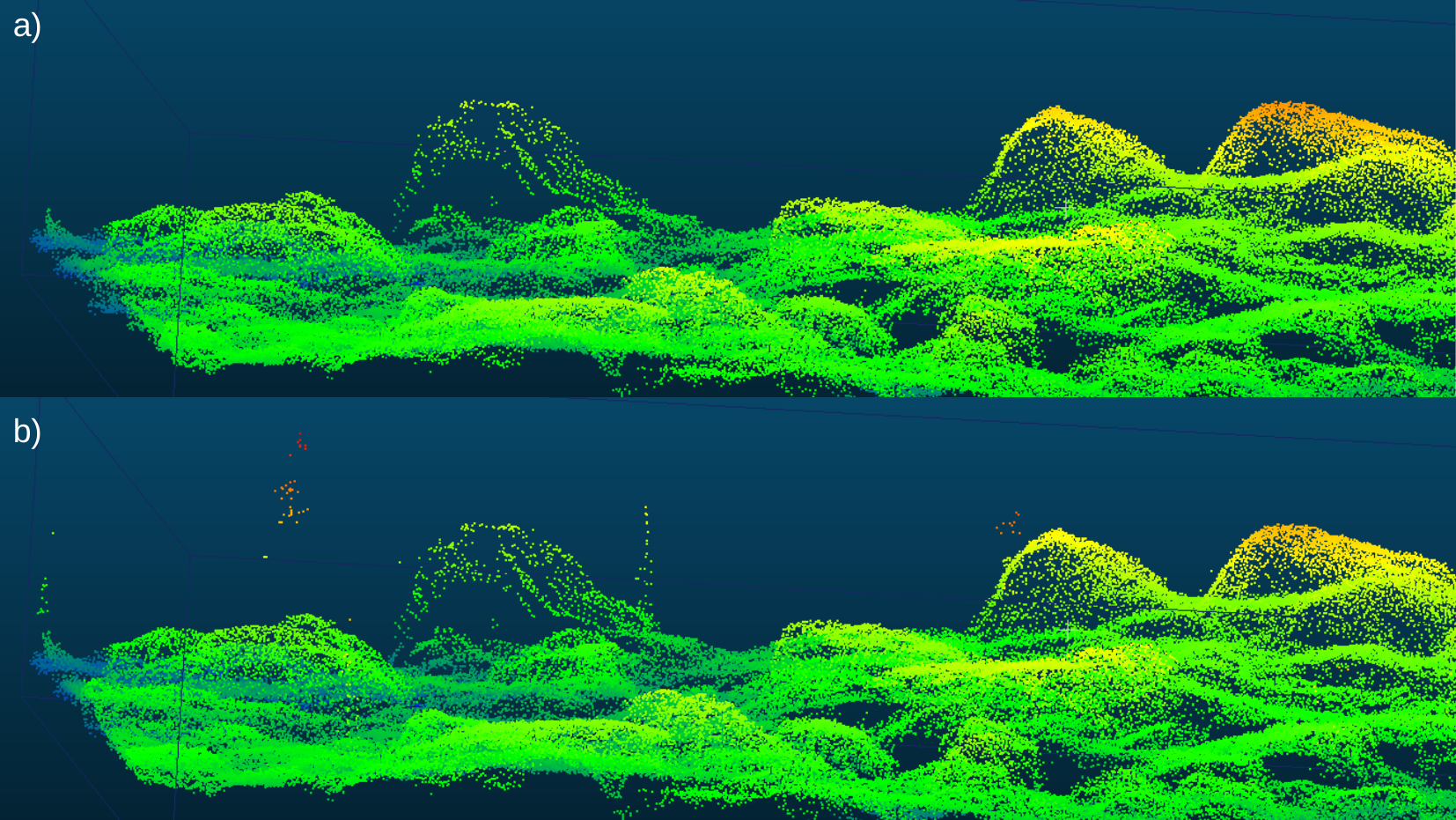}
    \caption{Example of a manual classification (a) and Terrascan classification (b) with some points misclassified as ground.}
    \label{fig:compare}
\end{figure}

\begin{figure}[h]
    \centering
    \includegraphics[width=0.6\textwidth]{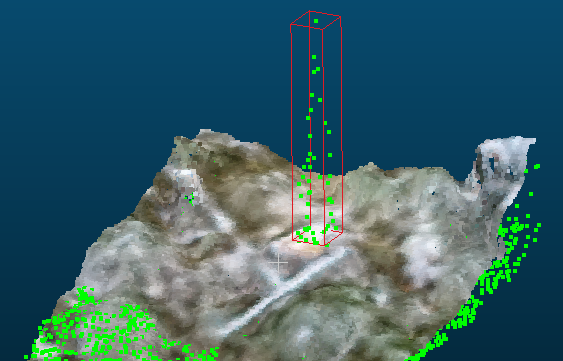}
    \caption{Example of a tree misclassified as ground (ground points above the stump).}
    \label{fig:tree}
\end{figure}

Figure \ref{fig:compare} shows an example of a manually annotated point cloud (top) compared to the initial classification made by Terrascan (bottom).
The manual classification is used as a reference for the evaluation of the classification algorithms.
An example of parts of a tree that is misclassified as ground is shown in Figure \ref{fig:tree}.

\subsection{Classification algorithms}
The classification algorithms used in this study are listed in Table \ref{tab:algos}.
Different approaches are used in the different software packages, such as morphological filters, TIN-based filters, and cloth simulation filters.
Cloth simulation filters (CSF) are used in three of the tested software packages (PDAL, LidR, and LiDAR 360) and is based on simulating a cloth being dropped over the inverted point cloud \citep{zhang2016easy}. Statistical analysis and processing scripts were mainly written in python and R, GUI-based software outputs were standardized and combined for evaluation and plotting in python.
For information about the logic behind the different algorithms, we refer to the references in Table \ref{tab:algos}.

\begin{table}[h] 
    \centering
    \caption{Algorithms for ground point classification}
    \label{tab:algos}
    \scriptsize
    \begin{tabular}[t]{p{2.5cm}p{2cm}p{3.5cm}p{4cm}}
      \hline
      Software & Algorithm & Reference & Comment \\
      \hline

      PDAL & CSF &  \citet{zhang2016easy} & \textbf{C}loth \textbf{S}imulation \textbf{F}ilter \\
      PDAL & PMF &  \citet{zhang2003progressive} & \textbf{P}rogressive \textbf{M}orphological \textbf{F}ilter \\
      PDAL & Skewness \newline balancing & \citet{bartels2010threshold} & Threshold-free classification \\
      PDAL & SMRF & \citet{pingel2013improved} & \textbf{S}imple \textbf{M}orphological \textbf{F}ilter \\
      \grayline
      LidR & PMF & \citet{zhang2003progressive} & \textbf{P}rogressive \textbf{M}orphological \textbf{F}ilter \\
      LidR & CSF &  \citet{zhang2016easy} & \textbf{C}loth \textbf{S}imulation \textbf{F}ilter \\
      LidR & MCC & \citet{evans2007multiscale} & \textbf{M}ultiscale \textbf{C}urvature \textbf{C}lassification \\
      \grayline
      Terrascan & TIN & \citet{axelsson2000generation} &  \\
      \grayline
      LiDAR 360 & IPTD & \citet{zhao2016improved} & \textbf{I}mproved \textbf{P}rogressive \textbf{T}IN \textbf{D}ensification \\
      LiDAR 360 & csf & \citet{zhang2016easy} & \textbf{C}loth \textbf{S}imulation \textbf{F}ilter \\
      \grayline
      Quick Terrain \newline Modeler & Classification \newline and extraction & - & Company-specific \\
      Quick Terrain \newline Modeler & Lidar Sense & - & Company-specific \\
      Quick Terrain \newline Modeler & QT Basic & - & Company-specific \\
      \grayline
      Lastools & TIN & - & based on \citet{axelsson2000generation} but modified \\
      \grayline
      Fusion & WLS & US Forest Service & \textbf{W}eighted Linear \textbf{L}east \textbf{S}quares  \\

      \hline
    \end{tabular}
\end{table}%

For each algorithm, relevant parameters are altered in order to investigate their effect on the classification results. The parameters are listed in Table \ref{tab:params}.
The tested parameter ranges were chosen based on the parameters available in each software package, vendor defaults, values suggested in the corresponding algorithm references when available, and a desire to cover both conservative and aggressive settings for obstacle-inclusive classification.
The ranges therefore differ between software packages, and the comparison should be interpreted as a benchmark of practically available workflows rather than an exhaustive parameter search for every implementation.

\begin{sidewaystable}[h]
    \centering
    \caption{Parameters for ground point classification algorithms}
    \label{tab:params}
    \scriptsize
    \begin{tabular}[t]{p{5cm}p{5cm}p{5cm}p{1.3cm}p{8cm}}
        \hline
        Software & Algorithm & Parameters & No. of settings & Parameter ranges \\
        \hline
        PDAL & CSF &  Resolution and rigidness & 15 & resolution: 0.1, 0.3, 0.5, 1.0, 2.0 \newline
                                                        rigidness: 1.0, 2.0, 3.0\\
        PDAL & PMF &  Max window, cell size, slope & 36 & max window: 4, 33, 66 \newline
                                                            cell size: 0.1, 0.5, 1.0 \newline
                                                            slope: 1.0, 5.0, 20, 50 \\
        PDAL & Skewness balancing & - & 1 & - \\
        PDAL & SMRF & Max window, cell size, slope, threshold & 96 & max window: 4, 18 \newline
                                                                    cell size: 0.1, 0.5, 1.0 \newline
                                                                    slope: 0.15, 0.25, 0.4, 0.6 \newline
                                                                    threshold: 0.5, 1.0, 2.0, 3.0 \\
        \grayline
        LidR & PMF & Max window & 4 & max window: 4, 12, 33, 66 \\
        LidR & CSF & Resolution, rigidness & 15 & resolution: 0.1, 0.3, 0.5, 1.0, 2.0 \newline
                                                    rigidness: 1.0, 2.0, 3.0 \\
        LidR & MCC & Scale, curve threshold & 25 & scale: 0.05, 0.1, 0.3, 0.6, 1.5 \newline
                                                    curve threshold: 0.01, 0.03, 0.06, 0.1, 0.3\\
        \grayline
        Terrascan & TIN & Iteration angle and distance & 2 & iteration angle: 6.0, 60 \newline
                                                                iteration distance: 1.4, 0.9 \\
        \grayline
        LiDAR 360 & IPTD & Scene, iteration angle and distance & 4 & scene: flat, gentle, hilly, steep \newline
                                                                    iteration angle: 6, 8, 10, 13 \newline
                                                                    iteration distance: 1.2, 1.4, 1.6, 1.7 \\
        LiDAR 360 & CSF & Scene, grid size & 3 & scene: flat, gentle, steep \newline
                                                grid size: 2, 1, 0.5 \\
        \grayline
        Quick Terrain Modeler & Classification \newline and extraction & - & 1 & - \\
        Quick Terrain Modeler & Lidar Sense & Grid sample & 2 & grid sample: 0.3, 2.0 \\
        Quick Terrain Modeler & QT Basic & Grid sample & 2 & grid sample: 0.3, 2.0 \\
        \grayline
        Lastools & TIN & Step, spike, bulge, offset & 3 & step: 5, 1.5 \newline
                                                            spike: nan, 2.0 \newline
                                                            bulge: 0.5, 2.0 \newline
                                                            offset: 0.05, 0.5 \\
        \grayline
        Fusion & WLS & g and w offset parameters & 6 & g offset: -2, -1, 0 \newline
                                                        w offset: 2.5, 4\\

        \hline
    \end{tabular}
\end{sidewaystable}%


\subsection{Evaluation}

To minimize the erroneous effects on field plot edges at raster generation, a 5 meter buffer around each plot is used.
For generating rasters, both the manual classification and the algorithm-derived classifications are rasterized using a bin-average method with a cell size of 0.25 m and 0.5 m for plots 1-5 and 6-7 respectively.
Raster cell size is set based on the ground point density in the manually classified LiDAR data.
Following best practice, each cell needs to include at least 4 ground points \citep{esri_best_dtm}.
The total point densities in Table \ref{tab:sites} gives ground point densities of between 27 and 620 points per square meter, resulting in an average number of ground points per raster cell of between 6.8 and 38.8 (Table \ref{tab:cellsize}).

\begin{table}[h]
    \centering
    \caption{Ground point densities after manual classification and raster cell sizes for the different field plots}
    \label{tab:cellsize}
    \scriptsize
    \begin{tabular}{p{1cm}p{1cm}p{2.8cm}p{1.8cm}p{2cm}}
        \hline
        \textbf{Site} & \textbf{Field plot} & \textbf{Point density (points/m$^2$)} & \textbf{Cell size (m)} & \textbf{Points per cell} \\
        \hline
        1 & 1 & 620 & 0.25 & 38.8 \\
        1 & 2 & 476 & 0.25 & 29.8 \\
        1 & 3 & 446 & 0.25 & 27.9 \\
        2 & 4 & 158 & 0.25 & 9.9 \\
        2 & 5 & 298 & 0.25 & 18.6 \\
        3 & 6 & 30 & 0.5 & 7.5 \\
        3 & 7 & 27 & 0.5 & 6.8 \\
        \hline
    \end{tabular}
\end{table}



Each raster cell was assigned the mean elevation of all classified ground points within that cell.
To obtain complete heightfields, a multiscale bin-averaging approach was used.
The point cloud was rasterized repeatedly at progressively finer resolutions, ranging from 16 m to 0.0625 m cell size.
Cells without points were initially left empty.
Starting from the coarsest raster, each subsequent finer raster was supplemented by values interpolated from the previous coarser level, but only in cells lacking observations.
Consequently, observed point elevations always took precedence, while larger-scale terrain information was used solely to fill gaps. This resulted in complete heightfields while preserving local terrain detail wherever ground points were available.
The manually classified point clouds, as well as their rasterized derivatives, are used as reference for the evaluation of the classification results.
Rasters without any interpolation are also produced, to be used in quantifying the plot coverage of each tested algorithm and setting.
Reported RMSE values are thus based on the interpolated rasters, while coverage is computed on rasters without interpolation to retain information about where a method failed to classify enough ground points.
Since coverage is computed by comparing the manually classified raster to the algorithm-derived raster, both without interpolation, the coverage metric can take values above 1 (100\% coverage) if the algorithm-derived raster was based on a point cloud with more ground points than the manually classified point cloud.
A coverage value of 1.1 is therefore interpreted as equally undesirable as a value of 0.9 within this metric, since both represent the same absolute deviation from complete one-to-one coverage.

The evaluation of the point cloud classification is performed as a point-by-point comparison to the manual classification using the \emph{$F_1$-score} (Eq. \ref{eq:f1score}),
\begin{equation}
  F_1\text{-score} = \frac{2 \cdot \text{Precision} \cdot \text{Recall}}{\text{Precision} + \text{Recall}},
  \label{eq:f1score}
\end{equation}
a harmonic mean of \emph{Precision} and \emph{Recall}, defined in Eq. \ref{eq:precision_and_recall} as
\begin{align}
    \text{Precision} &= \frac{TP}{TP + FP} \\
    \text{Recall} &= \frac{TP}{TP + FN}
    \label{eq:precision_and_recall}
\end{align}
where $TP$ is the number of true positives, $FP$ the number of false positives and $FN$ the number of false negatives.
We compare $F_1$-score with Matthews correlation coefficient (MCC; Eq. \ref{eq:mcc}),
\begin{equation}
    \text{MCC} = \frac{TP \cdot TN - FP \cdot FN}{\sqrt{(TP + FP)(TP + FN)(TN + FP)(TN + FN)}}
    \label{eq:mcc}
\end{equation}
where $TN$ is the number of true negatives, and also with Balanced Accuracy (BA; Eq. \ref{eq:ba}),
\begin{equation}
    \text{BA} = \frac{1}{2} \left( \frac{TP}{TP + FN} + \frac{TN}{TN + FP} \right)
    \label{eq:ba}
\end{equation}
to ensure robust evaluation \citep{chicco_matthews_2021}.
The evaluation of the rasters produced from the classified points is done using the Root Mean Square Error (RMSE; Eq. \ref{eq:rmse}),
\begin{equation}
    \text{RMSE} = \sqrt{\frac{1}{n} \sum_{i=1}^{n} (y_i - \hat{y}_i)^2}
    \label{eq:rmse}
\end{equation}
where $y_i$ is the raster cell value based on manual classification, $\hat{y}_i$ is the raster cell value based on algorithmic classification, and $n$ the number of cells in the raster.

\FloatBarrier

\section{Results}
The comparisons of results can be made per software, algorithm, and parameter settings, as well as for the different datasets/field plots.
Overall, a wide variety in performance is seen, with some algorithms performing well in some datasets and poorly in others, but also some algorithms performing well in all datasets.
The performance, in terms of e.g. $F_1$-score and RMSE, is often highly affected by the parameter settings used for each algorithm.

\subsection{Comparison of software and algorithms}
\begin{figure}[h]
    \centering
    \includegraphics[width=0.82\textwidth]{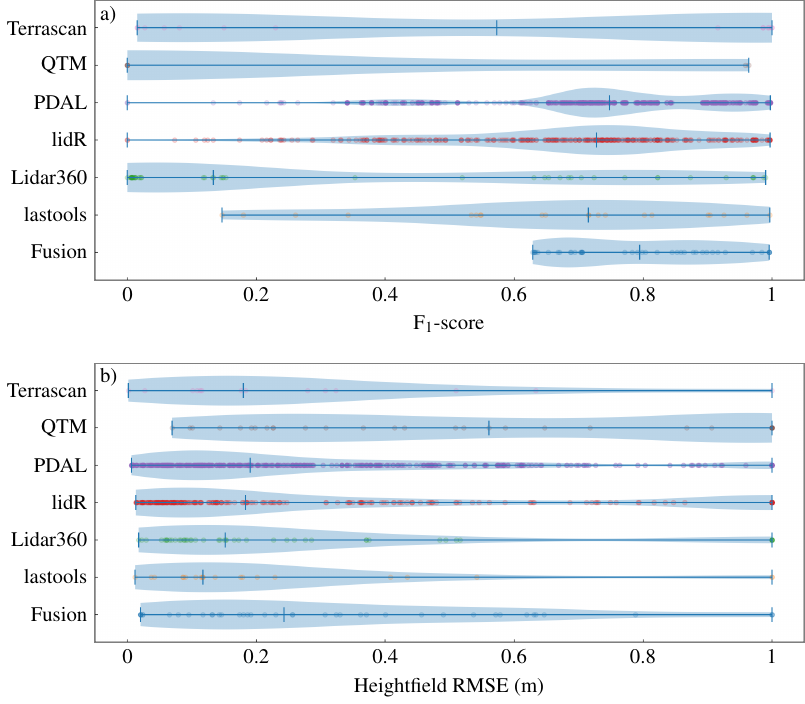}
    \caption{$F_1$-score (a) and raster/heightfield RMSE (b) for the different software. Individual algorithms, parameter settings, and field plots are shown as dots and the blue areas around lines describe data density. RMSE with values greater than 1 set to 1.}
    \label{fig:viol_f1_hf_av}
\end{figure}

\begin{figure}[h]
    \centering
    \includegraphics[width=0.82\textwidth]{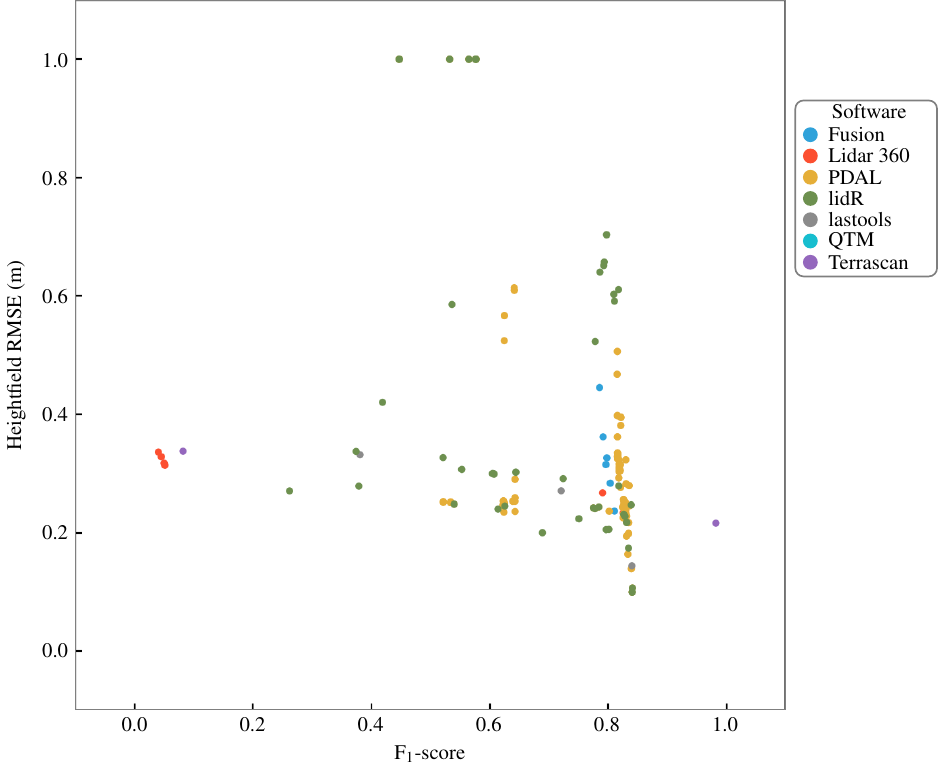}
    \caption{$F_1$-score and raster/heightfield RMSE, for the different software, average over field plots. Individual algorithms are shown as dots. RMSE values greater than 1 are set to 1.}
    \label{fig:f1_hf_av_soft}
\end{figure}

We first compare the software in terms of $F_1$-score for the point classification and RMSE of the rasters produced from the classified points.
Terrascan (with adapted settings, see Table \ref{tab:terrasettings}) stands out with a high $F_1$-score, on average as well as for each plot, see Figure \ref{fig:viol_f1_hf_av}.
This is expected since the manual classification is based on a Terrascan initialization.
For that reason, point-to-point metrics such as $F_1$-score, MCC, and BA should be interpreted with some caution, while raster RMSE is less directly tied to the initialization bias and is therefore a useful complementary comparison.
Interestingly, some methods, such as PDAL and lidR, achieved low raster RMSE despite lower point-wise scores than Terrascan.
Many of the results from PDAL, lidR, and Fusion are close to a $F_1$-score of 0.8, while having RMSE values ranging from 0.1 to 0.7 meters, compared to Terrascan with 0.2 meter RMSE value (Figure \ref{fig:f1_hf_av_soft}).

\FloatBarrier

\begin{figure}[h]
    \centering
    \includegraphics[width=0.95\textwidth]{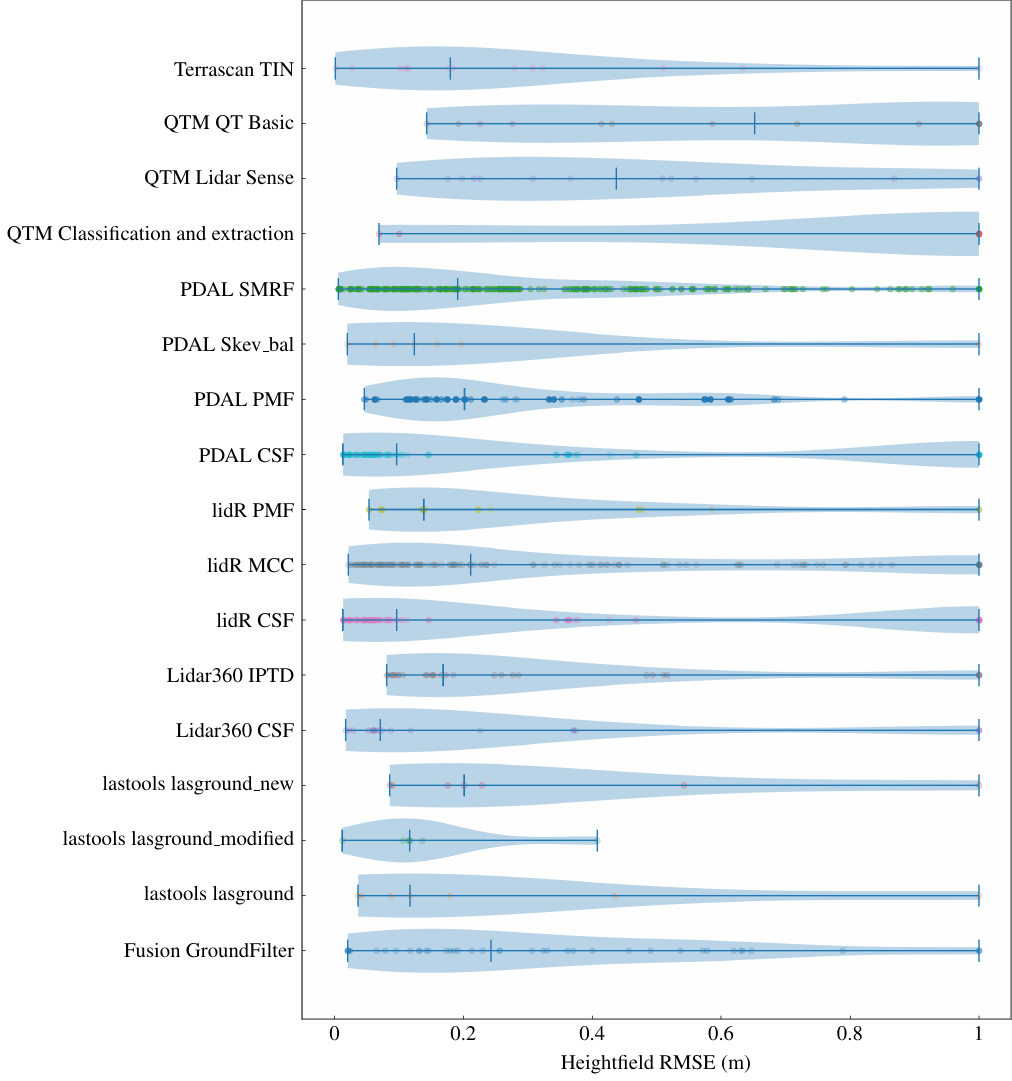}
    \caption{Raster/heightfield RMSE for the different software and algorithms. Individual field plots and parameter settings are shown as dots and the blue areas around lines describe data density. RMSE values greater than 1 are set to 1.}
    \label{fig:viol_both_hf}
\end{figure}

Combining software, algorithm, and the best performing parameter settings in terms of average raster RMSE over fieldplots, we see that the best results are obtained with LidR/PDAL CSF (Table \ref{tab:raster_diff}). 
When distinguishing the individual field plots, there are several algorithms that perform well on some field plots, but not on others.
Examples of this is Terrascan and Fusion (Table \ref{tab:raster_diff}, Figure \ref{fig:viol_both_hf}).
By average RMSE, the best configurations are PDAL/lidR CSF at 0.1 m cloth resolution and rigidness 3 (0.10), followed by PDAL SMRF and Lastools TIN (both 0.14), and lidR MCC (0.17).
At the same time, the spread across field plots remains substantial, especially on field plot 3, which indicates that average performance alone does not capture robustness across terrain conditions.

\begin{table}[h]
    \centering
    \caption{Raster difference (RMSE in meters) for the best performing software, algorithm, and parameter combinations across field plots}
    \label{tab:raster_diff}
    \scriptsize
    \begin{tabular}{p{5cm}p{1.1cm}p{0.7cm}p{0.7cm}p{0.7cm}p{0.7cm}p{0.7cm}p{0.7cm}p{0.7cm}}
        \hline
        \textbf{Software \& Algorithm} & \textbf{Average} & \textbf{Field plot 1} & \textbf{Field plot 2} & \textbf{Field plot 3} & \textbf{Field plot 4} & \textbf{Field plot 5} & \textbf{Field plot 6} & \textbf{Field plot 7} \\
        \hline
        PDAL CSF resol\_0.1 rigid\_3 & 0.10 & 0.08 & 0.05 & 0.36 & 0.06 & 0.04 & 0.08 & 0.01 \\
        PDAL CSF resol\_0.3 rigid\_3 & 0.26 & 0.34 & 0.05 & 1.29 & 0.04 & 0.02 & 0.06 & 0.02 \\
        PDAL PMF max-w\_33 cell-sz\_0.5 slope\_1 max-distance\_5 & 0.23 & 0.35 & 0.19 & 0.69 & 0.11 & 0.05 & 0.14 & 0.12 \\
        PDAL PMF max-w\_66 cell-sz\_0.1 slope\_5 max-distance\_5 & 0.24 & 0.33 & 0.20 & 0.58 & 0.14 & 0.13 & 0.16 & 0.12 \\
        PDAL SMRF max-w\_4 cell-sz\_0.5 slope\_0.6 threshold\_0.5 & 0.14 & 0.26 & 0.06 & 0.43 & 0.07 & 0.04 & 0.12 & 0.01 \\
        PDAL SMRF max-w\_18 cell-sz\_0.5 slope\_0.6 threshold\_0.5 & 0.14 & 0.26 & 0.06 & 0.43 & 0.07 & 0.04 & 0.12 & 0.01 \\
        PDAL skew-bal (fixed parameters) & 0.25 & 0.12 & 0.16 & 1.10 & 0.20 & 0.06 & 0.09 & 0.02 \\
        \grayline
        LidR CSF resol\_0.1 rigid\_3 & 0.10 & 0.08 & 0.05 & 0.36 & 0.06 & 0.04 & 0.08 & 0.01 \\
        LidR CSF resol\_0.3 rigid\_3 & 0.26 & 0.34 & 0.05 & 1.29 & 0.04 & 0.02 & 0.06 & 0.02 \\
        LidR MCC scale\_0.6 curve-threshold\_0.3 & 0.17 & 0.23 & 0.05 & 0.73 & 0.03 & 0.02 & 0.09 & 0.06 \\
        LidR MCC scale\_0.3 curve-threshold\_0.03 & 0.20 & 0.23 & 0.11 & 0.56 & 0.04 & 0.25 & 0.10 & 0.11 \\
        LidR PMF max-window\_4 & 0.24 & 0.47 & 0.22 & 0.59 & 0.05 & 0.07 & 0.14 & 0.14 \\
        LidR PMF max-window\_12 & 0.35 & 0.47 & 0.22 & 1.34 & 0.05 & 0.07 & 0.14 & 0.14 \\
        \grayline
        Terrascan TIN it-angle\_60 it-distance\_0.9 red-it-angle\_0.3 & 0.22 & 0.03 & 0.11 & 0.31 & 0.11 & 0.32 & 0.63 & 0.00 \\
        Terrascan TIN default-settings & 0.40 & 0.51 & 0.28 & 1.41 & 0.12 & 0.10 & 0.18 & 0.18 \\
        \grayline
        Lidar360 CSF SteepTerrain Grid\_0.5 ClassifyThr\_0.5 Iterations\_500 & 0.28 & 0.37 & 0.09 & 1.29 & 0.03 & 0.02 & 0.06 & 0.07 \\
        Lidar360 CSF GentleTerrain Grid\_1 ClassifyThr\_0.5 Iterations\_500 & 0.31 & 0.37 & 0.23 & 1.28 & 0.12 & 0.02 & 0.06 & 0.07 \\
        Lidar360 IPTD SteepTerrain Building\_20 Terrain\_88 IterationAngle\_13 IterationDist\_1.7 & 0.38 & 0.48 & 0.25 & 1.43 & 0.09 & 0.08 & 0.14 & 0.15 \\
        Lidar360 IPTD HillyTerrain Building\_20 Terrain\_88 IterationAngle\_10 IterationDist\_1.6 & 0.38 & 0.49 & 0.26 & 1.43 & 0.09 & 0.08 & 0.14 & 0.15 \\
        \grayline
        QTM ClassificationandExtraction gridsampling\_2 maxobjekt\_100 & 6.28 & 6.18 & 5.91 & 7.06 & 14.11 & 10.55 & 0.07 & 0.10 \\
        QTM LidarSense gridsampling\_0.3 maxobjektsize\_100 & 0.41 & 0.51 & 0.23 & 1.42 & 0.20 & 0.10 & 0.22 & 0.18 \\
        QTM QTBasic gridsampling\_2 & 0.56 & 0.59 & 0.72 & 1.25 & 0.43 & 0.23 & 0.28 & 0.41 \\
        QTM QTBasic gridsampling\_0.3 & 3.41 & 0.19 & 0.14 & 0.91 & 9.50 & 5.36 & 2.15 & 5.64 \\
        \grayline
        Lastools TIN lasground all-returns step\_1.5 spike\_2 bulge\_2 fine offset\_0.5 & 0.14 & 0.11 & 0.11 & 0.41 & 0.12 & 0.14 & 0.12 & 0.01 \\
        Lastools TIN lasground default-settings & 0.32 & 0.43 & 0.18 & 1.33 & 0.04 & 0.04 & 0.12 & 0.09 \\
        \grayline
        Fusion WLS g-par\_-2 w-par\_2.5 & 0.28 & 0.37 & 0.10 & 1.30 & 0.02 & 0.02 & 0.06 & 0.08 \\
        Fusion WLS g-par\_-1 w-par\_2.5 & 0.31 & 0.19 & 0.13 & 1.21 & 0.15 & 0.17 & 0.32 & 0.02 \\
        \hline
    \end{tabular}
\end{table}


\FloatBarrier

\begin{figure}[h]
    \centering
    \includegraphics[width=0.78\textwidth]{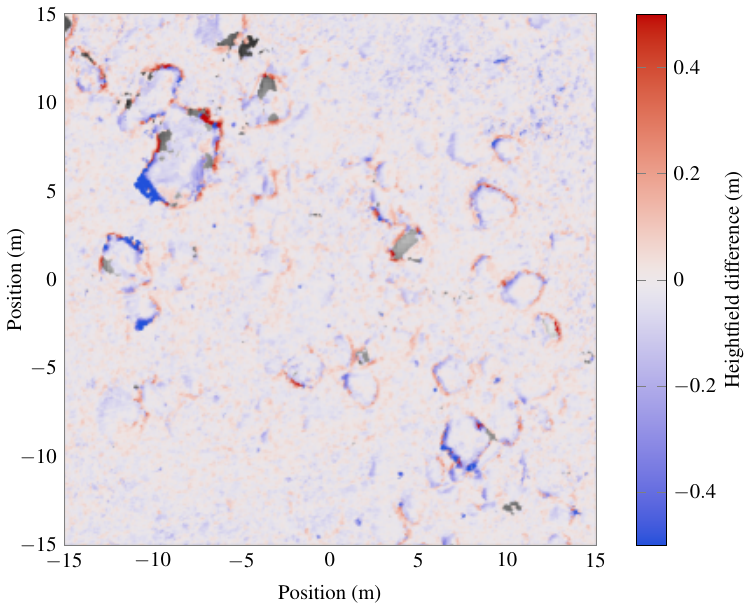}
    \caption{Difference generated by subtracting raster produced by manual classification from lidR CSF classification with cloth resolution 0.1 meter on field plot 1. The Precision measure is 0.56 and Recall 0.98.}
    \label{fig:lidR_CSF}
\end{figure}

\begin{figure}[h]
    \centering
    \includegraphics[width=0.7\textwidth]{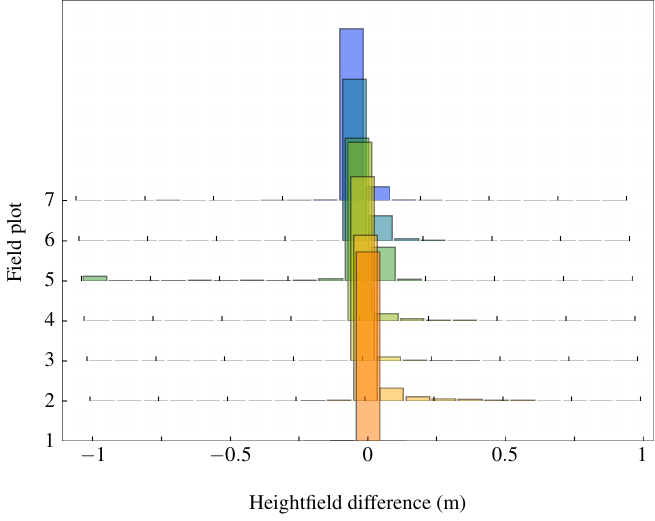}
    \caption{Histogram of differences generated by subtracting rasters produced by manual classification from lidR CSF classification, with cloth resolution 0.1 meter and rigidness 2, on respective field plots. Field plot 7 (top) in light blue.}
    \label{fig:csf_histo}
\end{figure}

An example of a raster produced from the manual classification and the classification from lidR CSF is shown in Figure \ref{fig:lidR_CSF} as the difference between the two rasters on field plot 1.
In this successful case, the algorithm has managed to classify most of the ground points correctly, including the boulders, and the RMSE after rasterization is 0.08 meters.
Differences appear at the edges between ground and boulders, where a small offset in the x or y direction results in a large difference in height.
Errors are slightly biased towards overestimation of the ground height, which is expected since the parameters are set to include boulders in the ground class (Figure \ref{fig:csf_histo}).
Although the precision is only 0.56, the recall is 0.98 and the raster RMSE is low.
For the present application, this tradeoff can still be favorable, since missing terrain obstacles is often more problematic than including some excess low vegetation, provided that the final terrain model remains geometrically close to the reference.
However, the areas with errors are small and the overall result is good.

\FloatBarrier

\subsection{The $F_1$ measure}
\begin{figure}[h]
    \centering
    \includegraphics[width=0.95\textwidth]{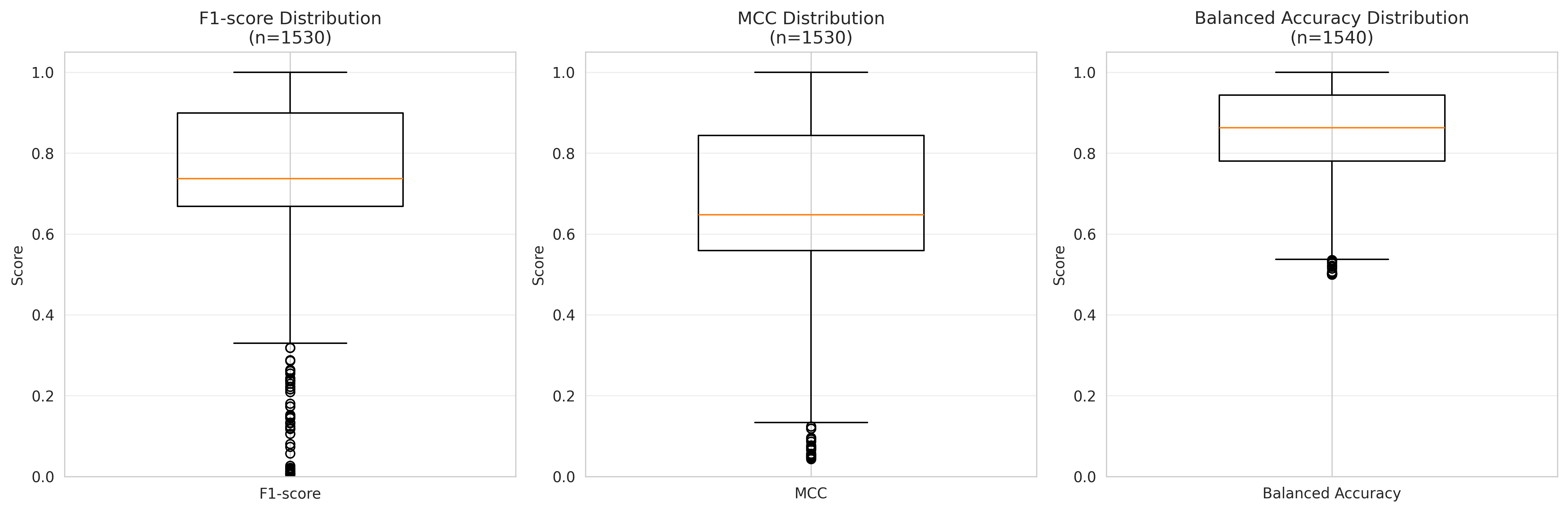}
    \caption{$F_1$-score, Matthew correlation coefficient (MCC) and Balanced Accuracy (BA) averaged for software, algorithms, parameter settings, and field plots.}
    \label{fig:f1_mcc_ba_overall}
\end{figure}

\begin{figure}[h]
    \centering
    \includegraphics[width=1.0\textwidth]{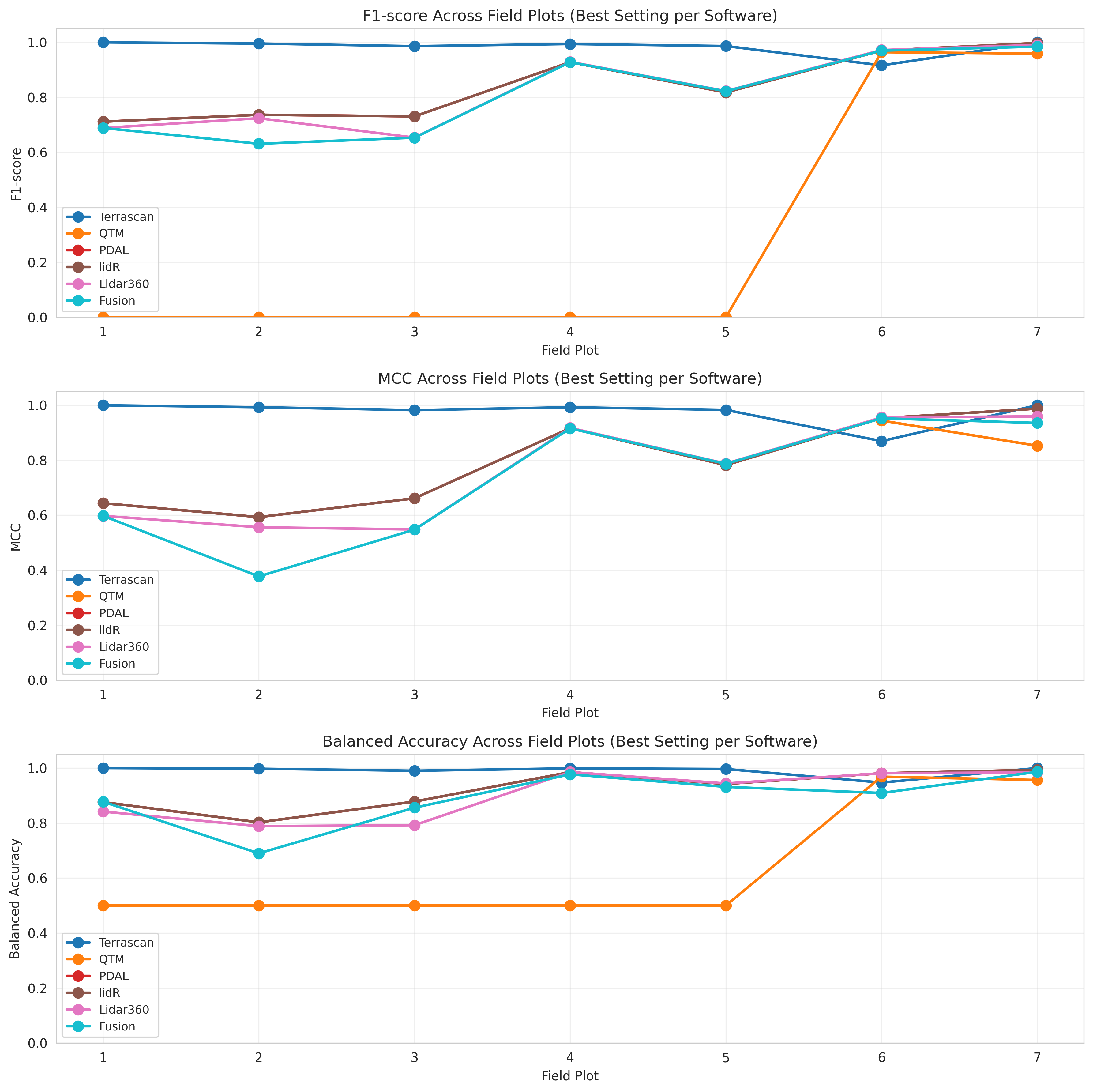}
    \caption{$F_1$-score, Matthew correlation coefficient (MCC) and Balanced Accuracy (BA) across field plots with best performing settings for each software. PDAL is identical to lidR (CSF with cloth resolution 0.1 meter and rigidness 1).}
    \label{fig:f1_mcc_ba_fieldplots}
\end{figure}

$F_1$-score only takes the 'positive' entries of the confusion matrix into account, and does not consider the 'negative' entries.
A certain algorithm or parameter setting might have a high $F_1$-score, but a bad performance in terms of true non-ground points over actual non-ground points.
The comparative analysis shows that the $F_1$-score and MCC measures are quite consistent, with BA showing a more liberal evaluation (Figures \ref{fig:f1_mcc_ba_overall} and \ref{fig:f1_mcc_ba_fieldplots}).
When separating the results per field plot and software, the best performing algorithms and parameter settings in terms of respective metric also confirm overall similarity between $F_1$-score and MCC (Figure \ref{fig:f1_mcc_ba_fieldplots}).
PDAL and lidR CSF with cloth resolution 0.1 meter and rigidness 1 show the exact same results in terms of $F_1$-score, MCC, and BA across all field plots, explaining the overlapping points in Figure \ref{fig:f1_mcc_ba_fieldplots}.

\subsection{Effect of terrain conditions}
\begin{figure}[h]
    \centering
    \includegraphics[width=0.9\textwidth]{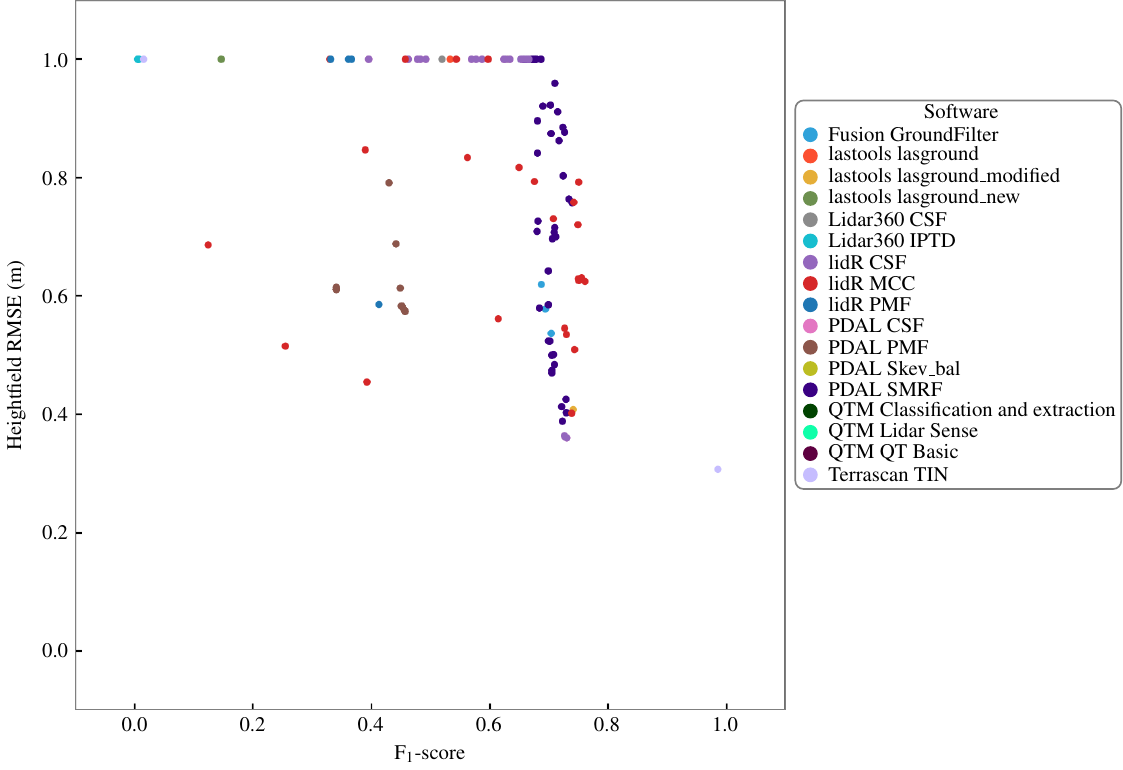}
    \caption{$F_1$-score and raster/heightfield RMSE for the different software and algorithms on field plot 3. RMSE values greater than 1 are set to 1.}
    \label{fig:plot_3}
\end{figure}

\begin{figure}[h]
    \centering
    \includegraphics[width=1.0\textwidth]{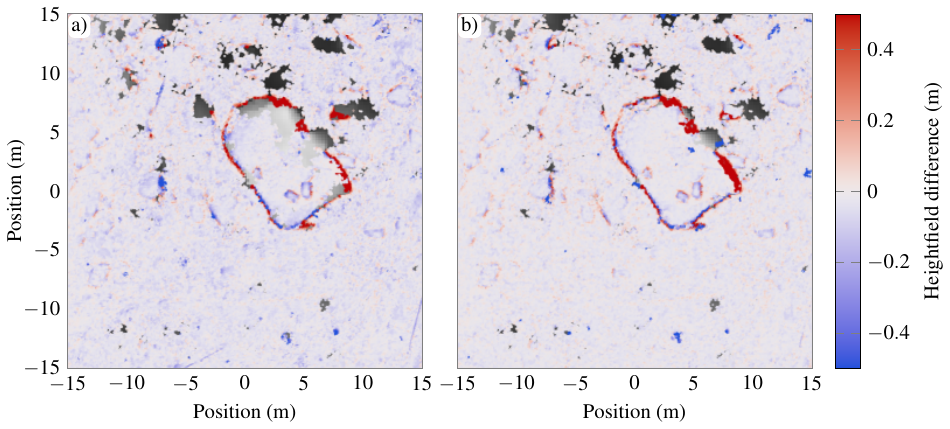}
    \caption{Differences between rasters of field plot 3 based on manual classification and PDAL-CSF (a) and Terrascan (b) ground point classification. The classification of Terrascan shows an $F_1$-score of 0.98 and an RMSE of 0.31 meters, while PDAL-CSF has a $F_1$-score of 0.73 and an RMSE of 0.36 meters.}
    \label{fig:diffs_plot_3}
\end{figure}

The studied field plots have different terrain conditions, ranging from very rough plots with big boulders to smooth plots with a slight slope.
Even though a good result is obtained for certain algorithms and parameter settings on average, the performance can be affected by certain geometric features and obstacles.
One example is field plot 3, which has a very large boulder of several meters size in all directions.
The algorithm that manages to classify this boulder as ground to the largest extent is Terrascan, see Figure \ref{fig:plot_3} and \ref{fig:diffs_plot_3}.
The second best algorithm is PDAL and lidR CSF, with an RMSE of 0.36 meters at 0.1 meters resolution and rigidness 2, compared to Terrascan's 0.31 meters, and it partly fails to follow the shape of the big boulder as well as leave slightly more area uncovered, visible as the grey heightmap instead of colored diff-values in Figure \ref{fig:diffs_plot_3}.
Uncovered areas in the raster are a result of no ground points being classified in those areas, leading to empty raster cells that need to be interpolated from surrounding cells with some method to achieve a complete DTM.
Under the tested settings, the software implementation of the CSF algorithm in PDAL and lidR did not affect the results.

\FloatBarrier

\begin{figure}[h]
    \centering
    \includegraphics[width=0.9\textwidth]{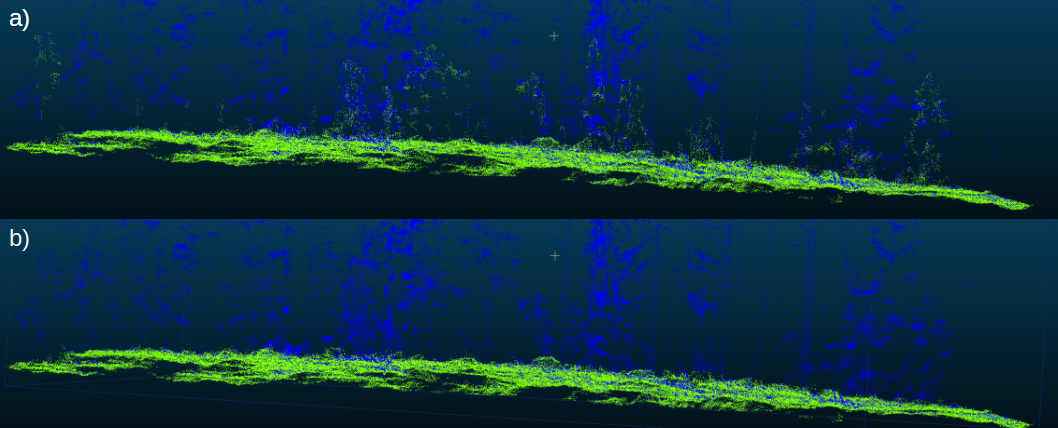}
    \caption{Terrascan classification (a) and manual classification (b) on field plot 5.}
    \label{fig:terrascan_manual_5}
\end{figure}

\begin{figure}[h]
    \centering
    \includegraphics[width=0.9\textwidth]{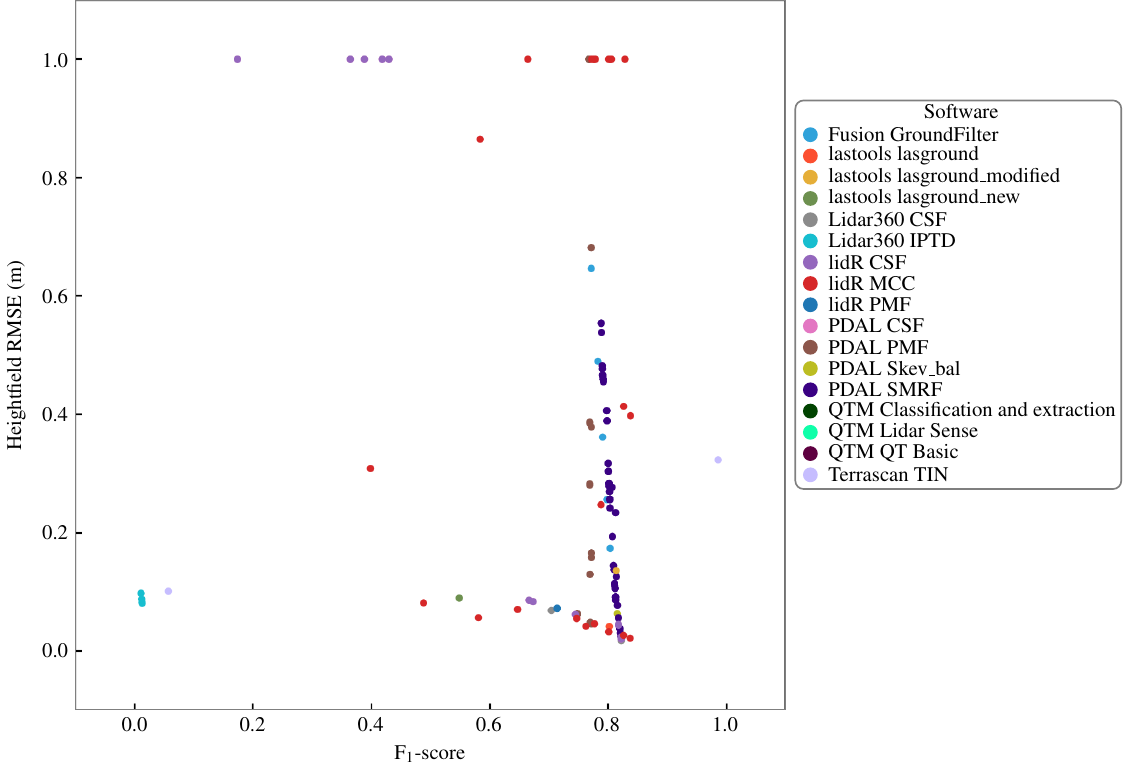}
    \caption{$F_1$-score and raster/heightfield RMSE for the different software and algorithms on field plot 5. RMSE values greater than 1 are set to 1.}
    \label{fig:plot_5}
\end{figure}

\begin{figure}[h]
    \centering
    \includegraphics[width=0.8\textwidth]{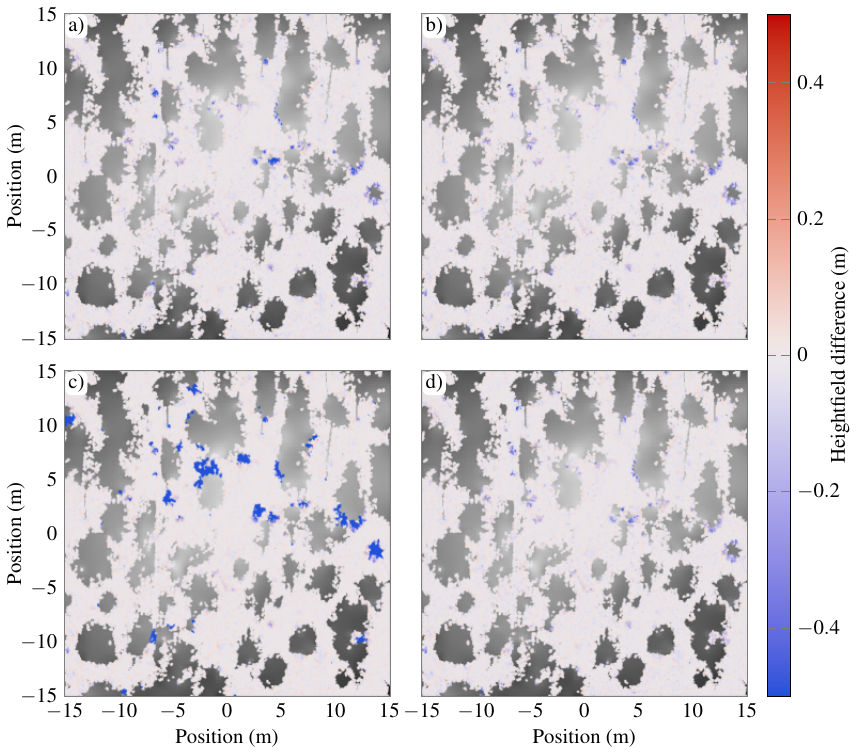}
    \caption{Differences between rasters of field plot 5 based on manual classification and PDAL-CSF 0.1 (a), PDAL-CSF 0.3 (b), Terrascan (c), and Fusion (d).}
    \label{fig:diffs_plot_5}
\end{figure}

Another terrain feature that affects the classification results is the presence of bushes and trees.
On field plot 5, Terrascan's aggressive settings (which show impressive results overall) include too much bushes and trees.
All the green points in Figure \ref{fig:terrascan_manual_5} appearing in the twigs and branches of the short trees and bushes are misclassified as ground.
Many of the other algorithms and settings obtain a better raster-to-raster RMSE on this particular field plot, even though their $F_1$-score is lower (Figure \ref{fig:plot_5}).
Comparing the generally successful CSF algorithm (0.1 meter resolution and rigidness 2) with Terrascan, we see that the CSF algorithm manages to classify more of the brush as non-ground, resulting in a lower RMSE (around 0.05 compared to 0.32 for Terrascan) (Figure \ref{fig:diffs_plot_5}).
What is interesting is that the finest resolution parameter setting for CSF (0.1 meters) also tends to capture more of the brush falsely as ground, while the next step (0.3 meters) performs better in this regard.
This suggests a recurring parameter tradeoff in the tested datasets: finer settings help preserve sharp obstacle geometry, but they also increase contamination from low vegetation.

\FloatBarrier

\subsection{Impact of point density}
\begin{figure}[h]
    \centering
    \includegraphics[width=0.9\textwidth]{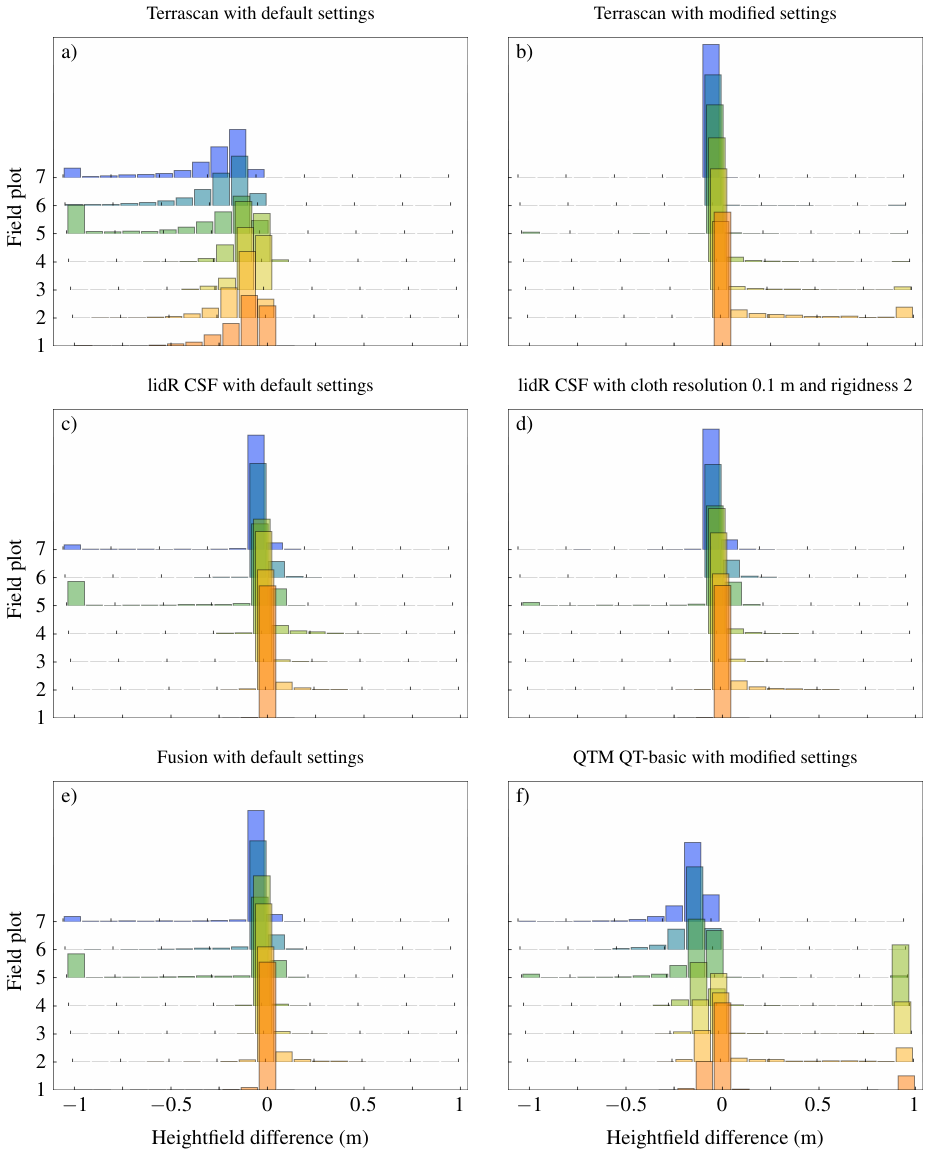}
    \caption{Histograms of differences generated by subtracting rasters produced by manual classification from algorithm generated classification on field plots 1 to 7.}
    \label{fig:histo_6}
\end{figure}

Field plots 1-3 have a high point density, while field plots 6-7 have a low point density.
This section should be interpreted as an exploratory comparison between scenarios with different point densities rather than as an isolated analysis of density alone, since density covaries with sensor type, acquisition geometry, and scene conditions in the available datasets.
When comparing field plots 1 and 7, with similar roughness but different point densities, the classification results show slightly more height differences in the high point density field plot 1 compared to field plot 7 (Figure \ref{fig:histo_6}).
On the other hand, the low point density in field plot 7 makes it unfeasible to use the same raster resolution as for field plot 1 (0.25 m cell size), which results in a coarser DTM with 0.5 m cell size (Table \ref{tab:cellsize}).
There are clear improvements in all field plots when using the modified settings for Terrascan and lidR CSF, while Fusion performs best with default settings and QTM struggles to get a good result at all (Figure \ref{fig:histo_6}).

\FloatBarrier

\subsection{Computing time}
\begin{figure}[h]
    \centering
    \includegraphics[width=0.82\textwidth]{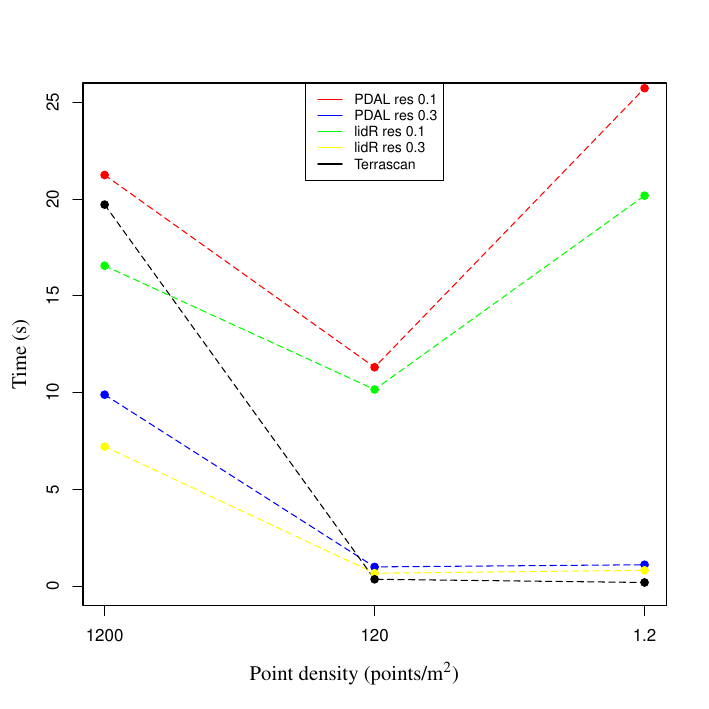}
    \caption{Computing time for point densities 1200 (original), 120, and 1.2 points/m$^2$ for a subset of the best performing algorithms and parameter settings. Results are shown as median of 10 runs.}
    \label{fig:time_point_density}
\end{figure}

The computing time for a subset of the best performing algorithms and parameter settings is compared for different point densities in Figure \ref{fig:time_point_density}.
The test is done on the same desktop workstation for all software, with an Intel Core i7-6700K CPU @ 4 GHz, 64 GB of RAM, and a 7200 RPM mechanical SATA hard disk.
A test area of one hectare is used for the computing time comparison, extracted from the same harvested area as field plots 1-3, a stony and rough area.
The average point density for the test area is 1200 points/m$^2$.
The computing time is measured for the automated processing chain from reading the point cloud from disk, through classification, to writing the classified point cloud to disk.
Each algorithm and parameter setting is run ten times, and the median time is used in the comparison.
Statistical analysis and processing script for the speed test was written in R.
Threading behavior and internal implementation details were not controlled uniformly across software packages. The timing comparison should therefore be interpreted as a practical end-to-end workflow comparison on this hardware configuration rather than as an absolute benchmark of algorithmic efficiency.
In this tested workflow and hardware configuration, the lidR CSF implementation is slightly faster than the PDAL CSF implementation while producing similar results in terms of $F_1$-score and RMSE.
Both software implementations of CSF are sensitive to fine cloth resolution settings in combination with a sparser point cloud, as seen in Figure \ref{fig:time_point_density}, where the computing time increases drastically when the cloth resolution is set to 0.1 meters.

\FloatBarrier

\section{Discussion}
Overall, the results from the different software, algorithms, and parameter settings show that existing ground-classification algorithms can be adapted to include boulders and other terrain obstacles relevant for traversability analysis.
Accuracy in terms of raster RMSE is in many cases substantially better than historical benchmarks for bare ground DTM generation in forested areas \citep{cateanu_als_2017}, although direct comparison is difficult due to differences in goal of the classification, point density, terrain, and evaluation methods.
A tradeoff between correctly classifying ground points including boulders and avoiding misclassifying non-ground points such as bushes and tree branches as ground points is evident in the results, specifically in field plot 5.
In practice, this is likely to be a common challenge, to a larger extent than is evident in this study, since bushes and small trees are common in forested areas.
The classification of undergrowth as opposed to bare ground as well as bigger trees have been challenging also in other studies \citep{puliti_for-instance_2023}.
More work is probably needed to develop a custom method that is more robust to different types of undergrowth and small trees, while still capturing boulders correctly.
In the tested datasets, this tradeoff appears to be at least as decisive as point density itself in determining practically useful parameter settings.

Several efforts to characterize forest terrain and obstacles for forest operations purposes have been made recently, focusing on different aspects of the problem and using different types of data and processing algorithms \citep{grube_automated_2025,hruza_application_2025}.
The current study adds to this body of emerging knowledge by providing a benchmark of existing ground point classification algorithms, enabling future studies to build upon these results with more confidence in the choice of software and algorithms to use as a baseline.

\subsection{Limitations}
A limitation of this study is that the manual classification used as ground truth is based on the best performing algorithm (Terrascan with modified settings), with manual corrections based on different support data for different field plots.
This might lead to a bias in the evaluation, favoring Terrascan over other software and algorithms, particularly for point-by-point metrics such as $F_1$-score, MCC, and BA.
There is also a risk for greater human error in the field plots where less support data was available, which also coincides with the field plots not visited by the authors (field plots 4-7).
Even though one big laser scanning campaign covering all field plots with their variation in geography, terrain type, and vegetation would have been ideal, such campaigns are costly and time-consuming and were not possible for this specific study.
Other studies sometimes analyse the effect of point density by downsampling the point cloud to different levels, while this study instead uses field plots with different original point densities, which means that other factors than point density, such as terrain type, vegetation, sensor properties, and acquisition geometry, might also affect the results.
On the other hand, downsampling is not always representative of real-world sparser ALS data, where the laser pulses might interact differently with the terrain and vegetation compared to a denser ALS acquisition.
With these limitations in mind, the results should still be valuable as a first benchmark and overview of how different software, algorithms, and parameter settings perform in this specific task of classifying ground points including boulders in forested areas.

\subsection{Applications and future work}
Future work includes developing methods for automatic segmentation of different types of obstacles, such as boulders, stumps and bushes, directly in point clouds, and in that way be able to spatially quantify the amount of obstacles in a forested area.
Applications of these highly detailed terrain models include, but are not limited to, terrain roughness estimation, vehicle-terrain simulations and trafficability analyses, more accurate soil moisture and soil bearing capacity estimations, and more precise choice of tree seedling plantation points.

\section{Conclusions}
We conclude that existing ground-classification algorithms can, under the tested conditions, be adapted to represent terrain obstacles relevant for vehicle traversability analysis by redefining the ground class and tuning parameters accordingly.
In terms of RMSE of the rasters produced from the classified points, the CSF algorithms in PDAL and lidR, with a cloth resolution of 0.1 m, perform best on the denser tested point clouds (field plots 1-3) while cloth resolution 0.3 m performs best on the sparser tested point clouds (field plots 4-7), suggesting that attention to point cloud density is important when selecting algorithm parameters.
Complex terrains with large presence and size variation of obstacles are, in the tested datasets, better classified with more aggressive or finer-resolution parameter settings, although this increases the risk of misclassifying non-ground points such as bushes and tree branches as ground points.
In the tested workflow and hardware configuration, lidR CSF is slightly faster than PDAL CSF, and the processing speed of both is sensitive to fine cloth resolution settings in sparse point clouds.


\section*{Acknowledgements}
The authors would like to thank the Swedish forest company SCA for providing the ALS data used in this study, as well as for running some of the algorithms. The forest owners' association Södra is acknowledged for providing the field inventory of rocks at Asa Research Park.

The authors utilized GPT-4o by OpenAI and Claude 4.5 by Anthropic for language improvement and coding assistance during the preparation of this manuscript. The output from these tools were reviewed and edited to ensure accuracy and alignment with the research objectives. All ideation remains of human origin.


\section*{Competing interests}
The authors declare no competing interests.

\section*{Funding}
This work was supported in part by Mistra Digital Forest (Grant DIA 2017/14 \#6) and United States Department of Agriculture, National Institute of Food and Agriculture, Agriculture and Food Research Initiative (USDA-NIFA-AFRI) under award number 20246701942511.

\section*{Data availability}
The data that support the findings of this study are available from the corresponding author upon request.

\section*{Author contributions}
ML, EW, SM, MS, MN and HP designed the study.
ML, SM, MN and HP performed data collection, analysis and interpretation for individual algorithms.
ML and EW performed the merging and overall interpretation of results, and the final evaluation analysis.
ML wrote the manuscript with contributions and revisions from all authors.
All authors read and approved the final manuscript and agree to be accountable for all aspects of the work.

\FloatBarrier
\bibliographystyle{unsrtnat}
\bibliography{references}

\end{document}